\documentclass[11pt,a4paper]{article}
\usepackage[hyperref]{acl2019}
\usepackage{times}
\usepackage{latexsym}

\usepackage{url}

\aclfinalcopy 

\usepackage[utf8]{inputenc} 
\usepackage[T1]{fontenc}    
\usepackage{hyperref}       
\usepackage{url}            
\usepackage{booktabs}       
\usepackage{amsfonts}       
\usepackage{nicefrac}       
\usepackage{microtype}      

\usepackage{graphicx}
\usepackage{amsmath}
\usepackage{amssymb}
\usepackage{amsthm}
\usepackage{adjustbox}
\usepackage{bm}
\usepackage{tikz}
\usepackage{lettrine}
\usepackage[titletoc]{appendix}
\usepackage[font=small]{caption}
\usepackage{arydshln}
\usepackage{bbm}
\usepackage{titlesec}

\newcommand{\eat}[1]{}
\newcommand{\nop}[1]{}

\title{\textsc{DeepCopy}: Grounded Response Generation with \\ Hierarchical Pointer Networks}

\author{
	Semih Yavuz\thanks{ \ \ Work done while interning at Google AI.} \\
    University of California, Santa Barbara \\
    {\tt syavuz@cs.ucsb.edu} \\\And
    Abhinav Rastogi \\
    Google AI \\
    {\tt abhirast@google.com} \\\AND
    Guan-Lin Chao \\
    Carnegie Mellon University \\
    {\tt guanlinchao@cmu.edu} \\\And
    Dilek Hakkani-T\"{u}r \\
    Amazon Alexa AI \\
  {\tt dilek@iee.org} \\}

\date{}

\begin{document}
    \maketitle
    \begin{abstract}
    Recent advances in neural sequence-to-sequence models have led to promising results for several language generation-based tasks, including dialogue response generation, summarization, and machine translation. However, these models are known to have several problems, especially in the context of chit-chat based dialogue systems: they tend to generate short and dull responses that are often too generic. Furthermore, these models do not ground conversational responses on knowledge and facts, resulting in turns that are not accurate, informative and engaging for the users. In this paper, we propose and experiment with a series of response generation models that aim to serve in the general scenario where in addition to the dialogue context, relevant unstructured external knowledge in the form of text is also assumed to be available for models to harness. Our proposed approach extends pointer-generator networks \cite{See17} by allowing the decoder to hierarchically attend and copy from external knowledge in addition to the dialogue context. We empirically show the effectiveness of the proposed model compared to several baselines including \cite{Ghazvininejad2018Grounded, Zhang18ConvAI2} through both automatic evaluation metrics and human evaluation on \textsc{ConvAi2} dataset.
    \end{abstract}
    
    \section{Introduction}
    Recently, deep neural networks have achieved state-of-the-art results in various tasks including computer vision, natural language and speech processing. Specifically, neural sequence-to-sequence models \citep{Sutskever14,Bahdanau15} have led to great progress in important downstream NLP tasks like text summarization \citep{Rush15, Nallapati16, See17, Tan17,Yavuz2018Calcs}, machine translation \citep{Cho14, Sutskever14, Luong15Attention, Bahdanau15}, and reading comprehension \citep{Xiong17Coattention}. However, achieving satisfactory performance on dialogue still remains an open problem. This is because dialogues can have multiple valid responses with varying semantic content. This is vastly different from the aforementioned tasks, where the generation is more conveniently and uniquely constrained by the input source.
    
    Although neural models appear to generate meaningful responses when trained with sufficiently large datasets in the chit-chat setting, such generic chit-chat models reveal several weaknesses that were reported by previous research \cite{SerbanEtAlHRED,VinyalsLe}. Most common problems include inconsistency in personality, dull and generic responses, and unawareness of long-term dialogue context.

    To alleviate these limitations, we turn our focus on a different problem setting for dialogue response generation where the model is provided a set of relevant textual facts (speaker persona descriptions) and is allowed to harness this knowledge when generating responses in a multi-turn dialogue. 
    To handle the personality inconsistency issue, we ground our dialogue generation model on external knowledge facts which are a list of persona descriptions in our application~\citep{Li2016Personal, Zhang18ConvAI2}.
    We explicitly use the dialogue history as memory for the model to condition on which potentially encourages a more natural conversation flow.
    Towards encouraging generation of more specific and appropriate responses while avoiding generic and dull ones, we use a hierarchical pointer network in our model such that it can copy content from two sources: current dialogue history and persona descriptions.
    
    In this work, we propose a novel and general architecture \textsc{DeepCopy} that extends the attentional sequence-to-sequence model with a hierarchical pointer network that enables the decoder to jointly attend and copy tokens from any of the facts available as external knowledge in addition to the dialogue context (encoder input). This is achieved entirely in an end-to-end fashion through factoring the whole copy mechanism into following three hierarchies/components:
    (i) a token-level attention mechanism over the dialogue context to determine the probability of copying a token from the dialogue context, (ii) A hierarchical pointer network to determine the probability of copying a token from each fact, and (iii) An inter-source meta attention over the input sources \textit{dialogue context} and \textit{external knowledge}, which combines the two copying probabilities. Using these components, a single copying probability distribution over the unique tokens appearing in the model input is computed exploiting the well-defined hierarchy among them. In addition, the model is equipped with a soft switch mechanism between \textit{copying} and \textit{generation} modes similar to \cite{See17}, which allows us to softly combine the \textit{copying probabilities} with the decoder's \textit{generation probabilities} over a fixed vocabulary into a final output probability distribution over an extended vocabulary. We empirically show the effectiveness of the proposed \textsc{DeepCopy} model compared to several baselines including \cite{Ghazvininejad2018Grounded, Zhang18ConvAI2} on \textsc{ConvAi2} challenge.
    
    \section{Related Work}
    \vspace*{-1ex}
    Earlier work on data-driven, end-to-end approaches to conversational response generation treated the task as statistical machine translation, where the goal is to generate a response given the previous dialogue turn ~\citep{RitterEtAl2011,VinyalsLe}. While these studies resulted in a paradigm change compared to earlier work, they do not include mechanisms to represent conversation context. To tackle this problem and have a better representation of conversation context as input to generation, ~\citep{SerbanEtAlHRED} proposed hierarchical recurrent encoder-decoder (HRED) networks. HRED combines two RNNs, one at the token level, modeling individual turns, and one at the dialogue level, inputting turn representations from the token-level RNNs. However, utterances generated by such neural response generation systems are often generic and contentless~\citep{VinyalsLe}. To improve the diversity and content of generated responses, HRED was later extended with a latent variable that aims to model the higher level aspects (such as topic) of the generated responses, resulting in the VHRED approach~\citep{SerbanEtAlVHRED}.
    
    Another challenge for dialogue response generation is the integration of knowledge into the generated responses. ~\citep{Liu2018Knowledge} extracted facts relevant to a dialogue from knowledge using string matching, named entity recognition and linking, found additional entities from knowledge that are most relevant to the facts by a neural similarity scorer, and used these as input context features for the dialogue generation RNN. ~\citep{Ghazvininejad2018Grounded} used end-to-end memory networks to base the generated responses on knowledge, where an attention over the knowledge relevant to the conversation context is estimated, and multiple knowledge representations are included as input during the decoding of responses. In this work, we use end-to-end memory networks as a baseline.
    
    Although much research has focused on response generation in a chit-chat setting, models trained on large datasets of human-human interactions of diverse speaker characteristics often tend to generate responses which are too vague and generic (common for most speakers) or inconsistent in personality (switching between different speakers' characteristics). Recently, \citep{Zhang18ConvAI2} presented the \textsc{ConvAi2} challenge containing persona descriptions and over 10K real human chit-chats where speakers were required to converse based on their assigned persona. \cite{Li2016Personal} learned speaker persona embeddings from a single-speaker setting (e.g. Twitter posts) or a speaker-address style (human-human conversations) to generate personalized responses given a single utterance input. Another related work \cite{Raghu18HypMn} applies hierarchical memory network for task oriented dialog problem. In this work, we compare our model with \citep{Zhang18ConvAI2} which uses a memory-augmented sequence-to-sequence response generator grounded on the dialogue history and persona.
    
    \setlength{\belowdisplayskip}{5pt} \setlength{\belowdisplayshortskip}{0pt}
    \setlength{\abovedisplayskip}{5pt} \setlength{\abovedisplayshortskip}{0pt}
    
    \vspace*{-1ex}
    \section{Model}
    In this section, we first set up the problem, and then briefly revisit the baseline models using memory networks \cite{Sukhbaatar14MemNet} and pointer-generator networks \citep{See17}. Subsequently, we introduce the proposed \textsc{DeepCopy} model with a hierarchical pointer network and our training process.
    
    \subsection{Problem Setup}
    Let $\mathbf{x}=(x_1, x_2, \ldots, x_n)$ denote the tokens in the dialogue history. The dialogue is accompanied by a set of $K$ relevant supporting facts, where $\mathbf{f}^{(i)}=(f^{(i)}_1, f^{(i)}_2, \ldots, f^{(i)}_{n_i})$ is the list of tokens in the $i$-th fact. Our goal is to generate the response as a sequence of tokens $\mathbf{y}=(y_1, y_2, \ldots, y_m)$ using the dialogue history and supporting facts. Note here that we are not interested in retrieval/ranking based models \cite{Weston18RetNRef} which rely on a set of candidate responses. Generative models are essential for this problem because we want to incorporate content from new facts during inference which may not be present in the training set. Hence, using a predefined set of candidates may not ensure high coverage.
    
    \subsection{Baseline Models}
    In this section, we describe several baseline response generation models including the ones from existing work \cite{Ghazvininejad2018Grounded, Zhang18ConvAI2} and the in-house ones we propose as additional baselines. 
    
    \subsubsection{Seq2Seq} \label{subsubsection:Seq2Seq}
    In a sequence-to-sequence model with attention \cite{Bahdanau15}, a sequence of input tokens is encoded using an LSTM encoder. At decoder step $t$, the decoder state $h_{t}$, a context vector $c_t$ and the previous decoder output $y_{t-1}$ are used together to output a distribution over a fixed vocabulary of tokens obtained from the training set using a non-linear function. The context vector $c_t$ is an attention-weighted combination of the encoder outputs. In the following baseline models, we use different features as inputs to the encoder. The underlying model remains the same.
    
    \noindent \textbf{\textsc{Seq2Seq + NoFact}.} Only the dialogue context tokens $\mathbf{x}$ are used as input to the encoder. \\
    \noindent \textbf{\textsc{Seq2Seq + BestFactContext}.} We select the fact $\mathbf{f}^{(c)}$ whose tokens have highest unigram \textit{tf-idf} similarity to the dialogue context tokens. $[\mathbf{x} || \mathbf{f}^{(c)}]$ is then used as input to the encoder, where $||$ denotes concatenation. \\
    \noindent \textbf{\textsc{Seq2Seq + BestFactResponse}.} We select the fact $\mathbf{f}^{(r)}$ whose tokens have highest unigram \textit{tf-idf} similarity to the ground truth response. $[\mathbf{x} || \mathbf{f}^{(r)}]$ is used as input to the encoder. The aim of this experiment is to have a better understanding of the effect of fact selection on response generation, since using the ground truth for fact selection is not fair.
    
    \vspace*{-1ex}
    \subsubsection{Memory Network}
    Our variations of Seq2Seq models described in Section \ref{subsubsection:Seq2Seq} incorporate facts by concatenating them to the dialogue context. Memory networks \cite{Ghazvininejad2018Grounded, Zhang18ConvAI2} are a more principled approach to incorporating external facts. Similar to \cite{Ghazvininejad2018Grounded}, we use a context encoder to embed the context tokens $\mathbf{x}$ and obtain a list of outputs and final hidden state $u \in \mathbb{R}^d$. As outlined in  \cite{Ghazvininejad2018Grounded}, a fact $\mathbf{f}^{(i)}$ is embedded into key and value vectors $k_i$ and $m_i$, respectively. A summary $o \in \mathbb{R}^d$ of facts is then computed as an attention weighted combination of $(m_1, m_2, \ldots, m_K)$ by conditioning on $u$ and $(k_1, k_2, \ldots, k_K)$. We then combine the two summaries into $\hat{u} = u + o$, and use it to initialize the decoder state. We report results on the following variants:
    
    \begin{figure*}[!t]
        \center
    	\includegraphics[width=\textwidth]{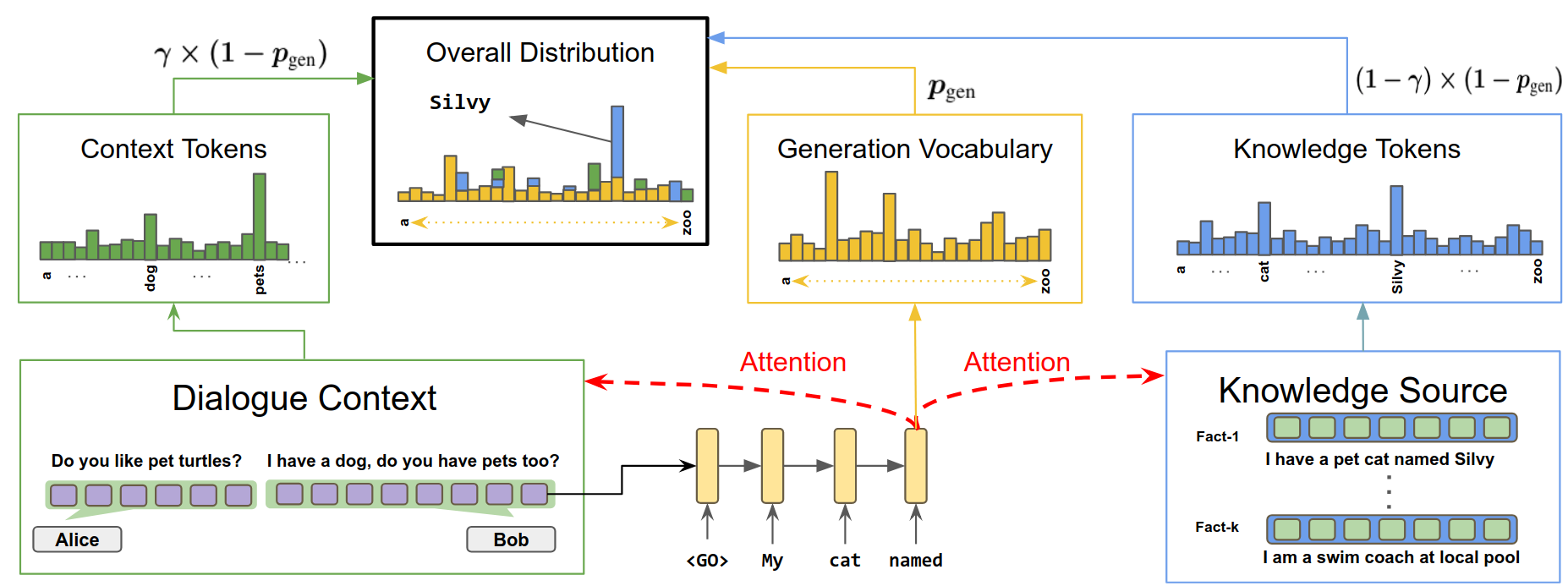}
    	\caption{Overview of our proposed approach as described in Section \ref{subsection:deep-copy}. The decoder state $d_t$ is used to attend over dialogue context and knowledge source to generate distributions for copying tokens from these sources. The decoder outputs a distribution over a fixed vocabulary. The three distributions are combined to yield the final distribution over tokens at each step $t$.}
    	\label{figure:deep_copy_overview}
    \end{figure*}
    
    \noindent \textbf{\textsc{MemNet}.} This is equivalent to the model used in \cite{Ghazvininejad2018Grounded}, described above. This is essentially a sequence to sequence model without attention at every decoder step, except using the combined summary $\hat{u}$ to initialize the decoder. \\
    \noindent \textbf{\textsc{MemNet+ContextAttention}.} At each decoder step, the decoder state attends over the encoder outputs and obtains a context vector $c_t^{(c)}$. This is equivalent to \textsc{Seq2Seq + NoFact} model from Section \ref{subsubsection:Seq2Seq}, except using the fact summary $\hat{u}$ to initialize the decoder state. \\
    \noindent \textbf{\textsc{MemNet+FactAttention}.} At each decoder step, we use the decoder state to attend over the value embeddings $(m_1, m_2, \ldots, m_K)$ corresponding to facts, and obtain a context vector $c_t^{(f)}$. This model is similar to the \textit{generative profile memory network} \cite{Zhang18ConvAI2}, where we apply attention only on facts, and we set the decoder's initial state to the combined summary $\hat{u}$. \\
    \noindent \textbf{\textsc{MemNet+FullAttention}.} This model employs attention over both facts and dialogue context at each decoder step. The two attention modules are combined by concatenating $c_t^{(c)}$ and $c_t^{(f)}$ \cite{zoph2016multi}.
    
    \subsubsection{Seq2Seq with Copy Mechanism}
    Seq2seq models can only generate tokens present in a fixed vocabulary obtained from the training set. Pointer-generator network \cite{See17} extends the attentional sequence-to-sequence model  \cite{Bahdanau15} by employing a pointer network \cite{Vinyals15Pointer}. It has two decoding modes, copying and generating, which are combined via a soft switch mechanism, allowing it to copy tokens from source in addition to generating from vocabulary.
    We report the results for the following additional baselines obtained by equipping the corresponding Seq2Seq model in Section \ref{subsubsection:Seq2Seq} with copy mechanism: \textsc{Seq2Seq + NoFact + Copy}, \textsc{Seq2Seq + BestFactContext + Copy}, \textsc{Seq2Seq + BestFactResponse + Copy}. 
    \nop{Note that we do not employ the coverage mechanism introduced by the original work \cite{See17} to prevent repetition for text summarization since it not as suitable for our task.}
    
    \begin{figure*}[!t]
        \center
    	\includegraphics[width=\textwidth]{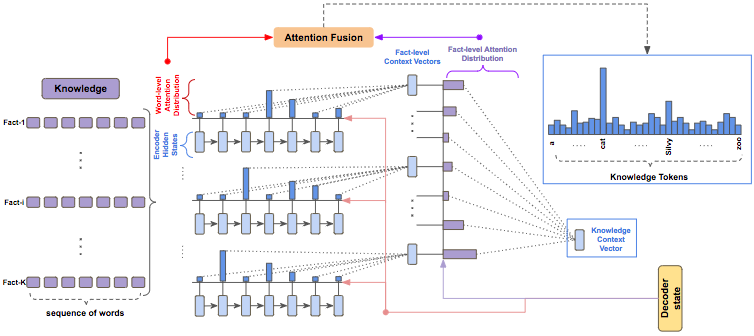}
    	\caption{Illustration of hierarchical pointer network. The decoder state $d_t$ is used to attend over tokens for each fact and also over the fact-level context vectors obtained by weighted average of token-level representations (w.r.t token-level attention weights) for each fact. The token-level attention weights are then combined with the attention distribution over facts (Equation \ref{eqn:fusion-att}) to generate the probability of copying each token in all the facts.}
    	\label{figure:deep_copy_intra_source}
    \end{figure*}
    \subsection{DeepCopy with Hierarchical Pointer Networks} \label{subsection:deep-copy}
    Pointer-generator network \cite{See17} can only copy tokens from the encoder input. In this section, we present our proposed \textsc{DeepCopy} model that extends pointer-generator network \cite{See17} using a novel hierarchical pointer network. Our model allows copying tokens from multiple input sources (facts $\mathbf{f}^{(i)}, 1 \leq i \leq K$), besides the encoder input (dialogue context $x$). 
    
    A high-level overview of the proposed approach is illustrated in Figure \ref{figure:deep_copy_overview}. At decoder step $t$, the decoder state $h_t$ is used to attend over the dialogue context tokens and fact tokens to give a distribution over the tokens present in context and facts respectively. These distributions are then combined with the distribution output by the decoder over the fixed vocabulary to obtain the overall distribution.\\
    \noindent \textbf{Encoding a sequence.~~} Let $\mathbf{w} = (w_1, w_2, \ldots w_n)$ be a sequence of tokens. We first obtain a trainable embedded representation of each token in the sequence and then use a LSTM cell to encode the sequence of embedding vectors. We define $e, \mathbf{s} = \texttt{Encode}(\mathbf{w})$, where $e$ denotes the final state of the LSTM and $\mathbf{s} = (s_1, s_2, \ldots s_n)$ denotes the outputs of the LSTM cell at all steps. \\
    \noindent \textbf{Attention.~~} Let $\mathbf{u} = (u_1, u_2, \ldots u_n)$ be a sequence of vectors where $u_i \in \mathbb{R}^p, 1 \leq i \leq n$ and $v \in \mathbb{R}^q$ be a conditioning vector. The attention module generates a linear combination $c$ of elements in $\mathbf{u}$ by conditioning them on $v$ as defined by the equations below. We define $\alpha, c = \texttt{Attention}(\mathbf{u}, v)$, where $\alpha_i \in \mathbb{R}^n$ is the weight assigned to $u_i$, and $c \in \mathbb{R}^p$ is a vector representation of the sequence $\mathbf{u}$ conditioned on $v$. In the equations below, $w_1$ and $W_2$ are parameters of appropriate dimension. In our setup, we use $p = q, w_1 \in \mathbb{R}^p$, and $W_2 \in \mathbb{R}^{p \times 2p}$.
    \begin{align}
        e_i &= w_{1}^T \tanh(W_2 [u_i; v]) \\
        \alpha_i &= \dfrac{\exp(e_i)}{\sum_{j=1}^n \exp(e_j)} \\
        c &= \sum_{i=1}^{n} \alpha_i u_i
    \end{align}
    \textbf{Copying from Dialogue Context.~~}
    Similar to our baseline models, we encode the dialogue context tokens $\mathbf{x}$ (Equation \ref{eqn:ctx-encode}) and apply attention to the encoder outputs at a decoder step $t$ (Equation \ref{eqn:ctx-attn}). This outputs attention weights $\alpha^{(x)}_t$ and a representation of the entire context $c^{(x)}_t$. The attention weights are aggregated to obtain the distribution over context tokens $p_t^{(x)}(w)$ (Equation \ref{eqn:ctx-copy}),
    \begin{align}
        \label{eqn:ctx-encode}
        e^{(x)}, \mathbf{s}^{(x)} &= \texttt{Encode}(\mathbf{x}) \\
        \label{eqn:ctx-attn}
        \alpha^{(x)}_t, c^{(x)}_t &= \texttt{Attention}(\mathbf{s}^{(x)}, h_t) \\
        \label{eqn:ctx-copy}
        p_t^{(x)}(w) &= \sum_{\{i: x_i = w\}} \alpha^{(x)}_{t, i}
    \end{align}
    \textbf{Copying from Facts: Hierarchical Pointer Network.~~}
    We introduce the hierarchical pointer network (Figure \ref{figure:deep_copy_intra_source}) as a general methodology for enabling token-level copy mechanism from multiple input sequences or facts. Each fact $\mathbf{f}^{(i)}$ is encoded (Equation \ref{eqn:facts-encode}) to obtain token level representations $\mathbf{s}^{(f)(i)}$ and overall representation $e^{(f)(i)}$. The decoder state $h_t$ is used to attend over token level representations (Equation \ref{eqn:facts-attn1}) and the overall fact-level representations of each fact (Equation \ref{eqn:facts-attn2}) by
    \begin{align}
        e^{(f)(i)}, \mathbf{s}^{(f)(i)} &= \texttt{Encode}(\mathbf{f}^{(i)}) \label{eqn:facts-encode} \\
        \alpha_{t}^{(f)(i)}, c_{t}^{(f)(i)} &= \texttt{Attention}(\mathbf{s}^{(f)(i)}, h_t) \label{eqn:facts-attn1} \\
        \beta_t, c^{(f)}_t &= \texttt{Attention}(\{c^{(f)(i)}_t\}_{i=1}^K, h_t) \label{eqn:facts-attn2}
    \end{align}
    to compute the probability of copying a word $w$ from facts as
    \begin{align}
        p^{(f)}_t(w) &= \sum_{j=1}^K \ p^{(f)}_t(\mathbf{f}^{(j)}) \cdot p^{(f)}_t(w |\mathbf{f}^{(j)}) \notag \\
                     &= \sum_{j=1}^K \ \beta_{t, j} \sum_{\{l: f^{(j)}_l=w\}} \alpha^{(f)(j)}_{t, l}
        \label{eqn:facts-copy}
    \end{align}
    
    \noindent
    \textbf{Inter-Source Attention Fusion~~}
    We now present the mechanism to fuse the two distributions $p_t^{(x)}(w)$ and $p_t^{(f)}(w)$ representing the probabilities of copying tokens from dialogue context and facts respectively. We use the decoder state $h_t$ to attend over dialogue context representation $c^{(x)}_t$ and overall fact representation $c^{(f)}_t$ (Equation \ref{eqn:fusion-att}). The resulting attention weight $\gamma_t' = [\gamma_t, 1 - \gamma_t]$ is used to combine the two copying distributions as shown in Equation \ref{eqn:fusion-prob}.
    \begin{align}
        \label{eqn:fusion-att}
        \gamma_t, c_t &= \texttt{Attention}([c^{(x)}_t, c^{(f)}_t], h_t) \\
        \label{eqn:fusion-prob}
        p^{\text{copy}}_t (w) &= \gamma_t \ p^{(x)}_t(w) + (1-\gamma_t) \ p^{(f)}_t(w)
    \end{align}
    Similar to Seq2Seq models, the decoder also outputs a distribution $p^{\text{vocab}}_t$ over the fixed training vocabulary at each decoder step using the overall context vector $c_t$ and decoder state $h_t$. Having defined the copy probabilities $p^{\text{copy}}_t$ for tokens that appear in the model input, either the dialogue context or the facts in external knowledge source, we combine $p^{\text{vocab}}_t$ and $p^{\text{copy}}_t$ using the mechanism outlined in \cite{See17}, except  we use $c_t$ defined in Equation \ref{eqn:fusion-att} as the context vector instead.
    
    To better isolate the effect of copying, a key component of the proposed \textsc{DeepCopy} model, we also conduct experiments with \textsc{MultiSeq2Seq} model that incorporates the knowledge facts in the same way (by encoding each fact separately with LSTM, and attending on each by the decoder as in \cite{zoph2016multi}), but relies completely on \textit{generation probabilities} without a copy mechanism.

    \subsection{Training}
    We train all the models described in this section using the same loss function optimization. More precisely, given a model $M$ that produces a probability $p_t(w | y_{<t})$ of generating token $w$ at decoding step $t$, we train the whole network end-to-end with the negative log-likelihood loss function of
    \begin{align*}
    J_{\text{loss}}(\mathbf{\Theta}) = - \dfrac{1}{|\mathbf{y}|} \sum_{t=1}^{|\mathbf{y}|}\log(p_t(y_t | y_{<t}, \mathbf{x}, \{\mathbf{f}^{(i)}\}_{i=1}^{K}))
    \end{align*}
    for a training sample $(\mathbf{x}, \mathbf{y}, \{\mathbf{f}^{(i)}\}_{i=1}^{K})$ where $\mathbf{\Theta}$ denotes all the learnable model parameters.
    
    \section{Experiments}
    In this section, we describe the details of dataset, training process, evaluation metrics, and the performance results of \textsc{DeepCopy} model in comparison to proposed and existing baselines.
    \vspace*{-1ex}
    \subsection{Dataset}
    We perform experiments for our problem setup on the recently released \textsc{ConvAi2} \textit{conversational AI challenge} dataset, which is an extended version of \textsc{PersonaChat} \cite{Zhang18ConvAI2}. The conversations in \textsc{ConvAi2} are obtained by asking a pair of crowdworkers to chat with each other naturally based on their randomly assigned personas (from a set of 1155 personas) towards getting to know each other. Personas are created by a different set of crowdworkers, and they consist of \textasciitilde 5 natural language sentences, each describing an aspect of a person that can range from common hobbies like \textit{"I like to play basketball"} to very specific facts like \textit{"I have a pet parrot named Tasha"}, reflecting a wide range of different personalities. The dataset contains \textasciitilde 11000 dialogues with \textasciitilde 160000 utterances, and 2000 dialogues with non-overlapping personas are used for validation and test. For our setting, we use personas as external knowledge sources that models can ground on while generating responses.
    \vspace*{-1ex}
    \subsection{Training and Implementation Details}
    In all the models explored in this paper, we set the dialogue context to concatenation of the last two dialogue turns separated by a special \texttt{CONCAT} token. The models are supplied with the persona facts of the side generating the response at the current turn, while the persona of the other side is concealed. We use a vocabulary of 18650 most frequent tokens and all the remaining tokens are replaced with a special \texttt{UNK} token. Embeddings of size 100 are randomly initialized and updated during training. We set the size of LSTM hidden layer to 100 for both encoder and decoder. The encoder and decoder vocabularies and embeddings are shared. A shared LSTM encoder is used for encoding both dialogue context and facts of external knowledge source. The model parameters are optimized using Adam \cite{Kingma15Adam} with a batch size of 32, a fixed learning rate of 0.001. We apply gradient clipping to 5 when its norm exceeds this value. During inference, we generate responses by employing a beam search of width 4. Our models are implemented in \textit{TensorFlow} \cite{Abadi16Tensorflow}.
    \vspace*{-1ex}
    \subsection{Main Results}
    In this section, we present the experimental results in terms of both automatic measures and human evaluation.
    \subsubsection{Automatic Evaluation}
    \begin{table*}[!t]
    	\centering
    	\begin{adjustbox}{max width=\textwidth}
    		\begin{tabular}{lccccc}
    			\hline  \\ [-2ex]
    			\textbf{Model} & \textbf{Perplexity} & \textbf{BLEU} & \textbf{ROUGE-L} & \textbf{CIDEr} & \textbf{Appropriateness} \\
    			\hline
    			\hline  \\ [-2ex]
                \textsc{[M-1] MemNet}  & 61.30 & 3.07 & 59.10 & 10.52 & 3.14 (0.51)\\
    			\textsc{[M-2] MemNet + ContextAttention} & 57.37 & 3.24 & 59.20 & 11.79 & 3.41 (0.54) \\
    			\textsc{[M-3] MemNet + FactAttention} & 61.50 & 2.43 & 59.34 & 9.65 & 1.45 (0.25)\\
    			\textsc{[M-4] MemNet + FullAttention} & 59.64 & 3.26 & 59.18 & 12.25 & 3.20 (0.49) \\
    			\hline  \\ [-2ex]
    			\textsc{[S2S-1] Seq2Seq + NoFact} & 60.48 & 3.38 & 59.46 & 11.41 & 3.12 (0.52) \\
    			\textsc{[S2S-2] Seq2Seq + BestFactContext} & 58.68 & 3.35 & 59.13 & 10.77 & 3.08 (0.45) \\
    			\textsc{[S2S-3] Seq2Seq + BestFactResponse*}  & 49.74 & 4.02 & 60.04 & 16.15 & 2.97 (0.51)\\
    			\hline  \\ [-2ex]
    			\textsc{[S2SC-1] Seq2Seq + NoFact + Copy} & 58.84 & 3.25 & 59.18 & 11.15 & 3.64 (0.54) \\
    			\textsc{[S2SC-2] Seq2Seq + BestFactContext + Copy} & 60.25 & 3.17 & 59.46 & 11.17 & 3.60 (0.51) \\
    			\textsc{[S2SC-3] Seq2Seq + BestFactResponse + Copy*} & 38.60 & 4.54 & 60.96 & 21.47 & 3.83 (0.46) \\
    			\hline  \\ [-2ex]
    			\textsc{[M-S2S] MultiSeq2Seq} (no \textsc{Copy}) & 57.94 & 2.88 & 59.10 & 10.92 & 3.32 (0.44) \\
    			$\textbf{\textsc{DeepCopy}}^{\dagger}$ & {\bf 54.58} & ${\bf 4.09}$ & {\bf 60.30} & {\bf 15.76} & {\bf 3.67} (0.59) \\
    			\hline  \\ [-2ex]
    			\textsc{G.Truth} & N/A & N/A & N/A & N/A & 4.40 (0.45) \\
    			\hline
    		\end{tabular}
    	\end{adjustbox}
    	\vspace{-2mm}
    	\caption[Table caption text]{Main results on \textsc{Convai2} dataset. Evaluation metrics on last three columns are better the higher. Perplexity is lower the better. The results of the proposed approach are presented in bold. * indicates that the corresponding model should be considered as a kind of \textbf{ORACLE} because it has access to the fact that is most relevant to the ground-truth response during the inference/test time as defined in Section \ref{subsubsection:Seq2Seq}. $\dagger$ indicates that the improvement of \textsc{DeepCopy} in automatic evaluation metrics over each of the other models (except \textsc{S2SC-3}) is statistically significant with p-value of less than 0.001 on the paired t-test.}
    	\label{table:convai_results}
    \end{table*}
    In Table \ref{table:convai_results}, we present our results in comparison with the existing and proposed baseline models. We report the performance of each model across several metrics commonly used for evaluation of text generation models including perplexity, corpus BLEU \cite{Papineni02Bleu}, ROUGE-L \cite{Lin04Rouge}, CIDEr \cite{Vedantam14Cider}.
    
    As expected, \textsc{Seq2Seq + BestFactResponse} model and its \textsc{+COPY} version outperform all the other models across all the evaluation metrics. This model pinpoints the importance of selecting the most suitable fact in the persona for the response to be generated at each turn, justifying our underlying motivation for conducting this experiment as highlighted in Section \ref{subsubsection:Seq2Seq}. However, the most suitable fact for the response is not available in the real application scenario, where the models are responsible for picking the useful pieces of information pertaining to the current dialogue turn to generate meaningful responses. Our proposed \textsc{Seq2Seq + BestFactContext} model and its \textsc{+COPY} version, on the other hand, are valid baselines for this scenario where the best fact is selected completely based on the dialogue context without relying on the ground-truth response. This model outperforms the previously proposed memory network based model \textsc{MemNet} \cite{Ghazvininejad2018Grounded} for knowledge grounded response generation on all the evaluation metrics, demonstrating its effectiveness despite the fact that it does not have access to all the facts unlike \cite{Ghazvininejad2018Grounded}. However, this approach has the following potential weaknesses: (i) if the best persona fact selected w.r.t dialogue context is wrong (irrelevant) for the ground-truth response, the generated response might be drastically misinforming, and furthermore it is difficult for model to recover from this error because it has no access to other facts, 
    (ii) selecting the best fact w.r.t dialogue context based on \textit{tf-idf} similarity may result in poor fact selection when the lexical overlap between context and response is small which might be a common case especially for the \textsc{ConvAi2} dataset as the focus of conversation may often change swiftly across the dialogue turns. The latter might be the reason why copying does not help much for this model since it might end up copying irrelevant tokens in the scenario mentioned above.
    
    \begin{table*}[!t]
    	\centering
    	\begin{adjustbox}{max width=\textwidth}
    		\begin{tabular}{lccccccc}
    			\hline  \\ [-2ex]
    			& \textbf{Diversity} & & & \textbf{Fact-Inclusion} & & & \textbf{Agreement}  \\
    			\hline
    			\textbf{Model} & \textbf{Distinct}-2 / 3 / 4 & & \textbf{F.Inc} & \textbf{F.Per} & \textbf{F.Hal} & & \textbf{F.Inc} / \textbf{F.Per}\\
    			\hline 
    			\hline  \\ [-2ex]
    			\textsc{M-1} & .004 / .006 / .010 & & 0.41 & 0.01 & 0.40 & & 0.99 / 0.99 \\
    			\textsc{M-2} & .010 / .019 / .031 & & 0.43 & 0.01 & 0.42 & & 0.97 / 0.99 \\
    			\textsc{M-3} & .001 / .001 / .002 & & 0.06 & 0.04 & 0.02 & & 0.99 / 0.99 \\
    			\textsc{M-4} & .054 / .010 / .156 & & 0.51 & 0.09 & 0.42 & & 0.98 / 0.98 \\
    			\hline  \\ [-2ex]
    			\textsc{S2S-1} & .012 / .022 / .036 & & N/A & N/A & N/A & & N/A / N/A \\
    			\textsc{S2S-2} & .012 / .022 / .035 & & 0.54 & 0.04 & 0.50 & & 0.97 / 0.99 \\
    			\textsc{S2S-3} & .026 / .043 / .061 & & 0.79 & 0.16 & 0.63 & & 0.97 / 0.97 \\
    			\hline  \\ [-2ex]
    			\textsc{S2SC-1} & .039 / .069 / .104 & &  N/A & N/A & N/A & & N/A / N/A \\
    			\textsc{S2SC-2} & .035 / .067 / .109 & & 0.73 & 0.36 & 0.37 & & 0.99 / 0.99\\
    			\textsc{S2SC-3*} & .058 / .111 / .178 & & 0.73 & 0.55 & 0.18 & & 0.98 / 0.96 \\
    			\hline  \\ [-2ex]
    			\textsc{M-S2S} & .035 / .065 / .104 & & 0.47 & 0.05 & 0.42 & & 0.96 / 0.98 \\
    			\textbf{\textsc{DeepCopy}} & \textbf{.059 / .121 / .201} & & 0.62 & 0.23 & 0.39 & & 0.95 / 0.97 \\
    			\hline  \\ [-2ex]
    			\textsc{G.Truth} & 0.35 / 0.66 / 0.84 & & 0.76 & 0.49 & 0.27 & & 0.93 / 0.96  \\
    			\hline
    		\end{tabular}
    	\end{adjustbox}
    	\caption[Table caption text]{Lexical diversity and fact inclusion analysis results. Model names are abbreviated according to Table \ref{table:convai_results}. \textbf{F.Inc} denotes the ratio of responses that include factual information. \textbf{F.Per} and \textbf{F.Hal} denote the ratio of responses where the included fact is consistent with the persona or a hallucinated one, respectively. \textbf{Agreement} column corresponds to Cohen's $\kappa$ statistic measuring inter-rater agreement on binary factual evaluation metrics for \textbf{F.Inc} and \textbf{F.Per}. * indicates the \textbf{ORACLE} model.}
    	\vspace{-1mm}
    	\label{table:fact_diversity_analysis}
    \end{table*}
    
    Our proposed \textsc{DeepCopy} model is designed to effectively address the aforementioned issues, where it has access to the entire set of persona facts per dialogue from which it is expected to include the useful pieces of information in the response. \textsc{DeepCopy} model outperforms all the models reported in Table \ref{table:convai_results} except for \textsc{Seq2Seq + BestContextResponse} models, which we already deem as kind of an upper bound because it has access to the most relevant fact to the response. This justifies the effectiveness of \textsc{DeepCopy} model compared to the existing works \cite{Ghazvininejad2018Grounded, Zhang18ConvAI2} and the additional baselines we explored in this work. On the other hand, \textsc{MultiSeq2Seq} performs considerably worse than the \textsc{DeepCopy} model despite the fact they both have access to the entire set of facts and employ the same encoder-decoder architecture except for the copy mechanism. This further justifies the effectiveness of incorporating the proposed hierarchical pointer networks in \textsc{DeepCopy} because integrating the external knowledge simply by employing multi-source attention as in \cite{zoph2016multi} does not yield to a good solution with competitive results, performing even worse than \textsc{Seq2Seq + NOFACT} on 3 of the metrics. 
    
    \subsubsection{Human Evaluation}
    Although automatic metrics provide tangible information regarding the performance of the models, we augment them with human evaluations for a more comprehensive analysis of the resulting model generated responses. Towards this end, we randomly sample 100 examples from test data and ask human raters to evaluate the candidate model generated responses in terms of appropriateness. Each example is rated by 3 raters, who are shown a dialog history along with a set of persona facts (of the person in turn), and asked to rate each response based on its \textit{appropriateness} in the dialogue context with a score from 1 (worst) to 5 (best).
    
    In Table \ref{table:convai_results}, we present the results of human evaluation under the \textit{appropriateness} column. Since each response is rated by $3$ different human raters, we report the average rating along with the standard deviation in parenthesis. We observe that \textsc{DeepCopy} outperforms both the existing memory-network baselines and the proposed sequence-to-sequence baselines on the appropriateness evaluation. It also achieves a performance that is close to the \textit{oracle} model (\textsc{S2SC-3}), which has a leverage of having an access to the fact that is most relevant to the ground-truth response during the inference time. Overall, human evaluation of the responses in terms of appropriateness further justifies the promise and effectiveness of our proposed \textsc{DeepCopy} model.
    
    Appropriateness scores also demonstrate the advantage of incorporating the soft copy mechanism. Comparing \textsc{S2S} (and \textsc{M-S2S}) models to their copy-equipped counterparts (\textsc{S2SC}) (and \textsc{DeepCopy}) in Table \ref{table:convai_results} immediately reveals a significant gain in appropriateness score. Another significant observation to note here is that ground-truth responses obtain an average appropriateness score of $4.4 / 5$, which reflects both the noise in \textsc{ConvAI2} dataset and the difficulty of generating the perfect response even for humans.
    
    \subsection{Further Analysis and Discussion}
    \noindent\textbf{Lexical Diversity Analysis}.
    In Table \ref{table:fact_diversity_analysis}, we report the lexical diversity results using the distinctness metric introduced in \cite{Jiwei2017Diversity}. \textit{distinct}-$n$ score corresponds to the number of distinct $n$-grams divided by total number of generated $n$-grams. We can clearly observe that \textsc{DeepCopy} generates the most diverse responses among all the models including the copy-augmented oracle model (\textsc{S2SC-3}). Hence, diversity results further show that our proposed model is promising in addressing the most commonly observed \textit{generic response problem} more effectively than existing models by generating more diverse responses.
    
    \noindent\textbf{Fact Inclusion Analysis}.
    We also conduct an analysis on the kinds of factual information included in the model-generated responses. More precisely, our goal is to understand how often the generated response includes a factual information (F.Inc), and whether this information is consistent with the persona facts (F.Per) or a hallucinated one (F.Hal). A good model can naturally include available facts from the persona and hallucinate others when the conversation context requires them.
    Towards this end, we ask $3$ human raters to label responses with 1 (or 0) based on whether a fact is included, and if so, whether this fact is a persona-fact or not.
    
    In Table \ref{table:fact_diversity_analysis}, we present an analysis for the kinds of factual information included in model generated responses. As can be seen from this analysis, models that have a copy mechanism include more facts from the persona than the ones that do not. Another important observation is that the ground-truth responses include facts from persona only in 49\% of the times, which indicates that the provided persona facts remain insufficient to cover the complexity of the high entropy open-ended person-to-person conversations.
    
    In Table \ref{table:fact_diversity_analysis}, we present Cohen's $\kappa$ score for each model and fact analysis metric pair using the scores from 3 raters for each example. We observe for each model and metric pair a $\kappa$ statistic of greater than 0.9, which indicates a near perfect agreement among raters. Note that the ratio of hallucinated facts (\textbf{F.Hal}) is derived directly from human labels for fact inclusion (\textbf{F.Inc}) and persona-fact (\textbf{F.Per}). That is why, there is no separate labelling process for hallucinated facts (\textbf{F.Hal}). Hence, there is no $\kappa$ statistic for \textbf{F.Hal} in Table \ref{table:fact_diversity_analysis}.
    
    \noindent{\textbf{Error Analysis}}.
    A deeper analysis of the examples where \textsc{DeepCopy} is assigned a worse appropriateness score than the best performing memory-network based baselines (\textsc{M-2} and \textsc{M-4}) reveals the following further insights: (i) Some of these examples are corresponding to the cases where a generic response (e.g., "I've a dog named radar", one of the frequent generic responses, completely independent of persona facts) is rated much higher (5 to 1) than factual but slightly off (by a single word in this example) responses (e.g., "I have a dog for a living." coming from the persona fact "I walk dogs for a living."), (ii) In another subset of the analyzed examples, \textsc{DeepCopy} model generates a response (e.g., "yes, but I want to become a lawyer.") by incorporating a fact that has already been used in the previous turn of the dialog whereas \textsc{M-2} produces a generic response (e.g., "that's great. do you have any hobbies?", again irrelevant to facts) which is rated higher.
    (iii) And most of the remaining cases fall into the class of examples where incorporating knowledge facts breaks the conversation flow, which is a crucial observation specific to this dataset that can also be supported by the low persona-fact inclusion ratio (49\%) of ground-truth responses.
    
    \subsection{Qualitative Observations}
    \begin{figure}[!t]
        \center
    	\includegraphics[width=0.5\textwidth]{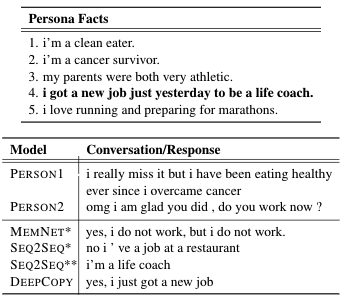}
    	\caption{Example dialogue where the previous two turns from \textsc{Person1} and \textsc{Person2} along with the responses generated by the models acting as \textsc{Person1} are shown on the right. Persona facts for \textsc{Person1} are provided on the left, among which the one in bold is the best fact w.r.t  response. \textsc{MemNet}*, \textsc{Seq2Seq}*, \textsc{Seq2Seq}** are abbreviations for \textsc{MemNet + FullAttention}, \textsc{Seq2Seq + BestFactResponse}, \textsc{Seq2Seq + BestFactResponse +COPY} models, respectively.}
    	\label{figure:dialogue_examples}
    \end{figure}
    
    In Figure \ref{figure:dialogue_examples}, we present an example dialogue where \textsc{DeepCopy} model generates a meaningful and fluent response by effectively mixing \textit{copy} and \textit{generate} modes. We can observe that it is able to attend on the right persona fact by taking the dialogue context (especially the question at the end of \textsc{Person2}'s turn) into consideration. Furthermore, attending to the tokens of this fact, it produces a fluent and valid answer to yes/no question by generating "yes" and copying the rest (and most) of the tokens from the fact. 
    Although it copies most of the tokens from the fact, it is good to observe that it copies exactly the relevant pieces instead of just copying the entire fact. 
    \textsc{Seq2Seq + BestFactResponse + Copy} model's response is also meaningful and fluent although it may not be as engaging for the continuation of dialog. However, the quality of the response by \textsc{Seq2Seq + BestFactResponse} quickly degrades compared to its \textsc{+Copy} version. Although the response is still fluent and relevant to the dialogue context, it becomes rather irrelevant to the persona as the model seems to have difficulty of picking the useful information from even the best persona fact it is provided with when the copy mechanism is disabled. Lastly, the response generated by \textsc{MemNet+FullAttention} model seems to still suffer from repetition, semantic consistency, and relevancy problems that were observed and reported by previous work.
    
    \vspace*{-1ex}
    \section{Conclusion and Future Work}
    \vspace*{-1ex}
    We propose a hierarchical pointer network for knowledge grounded dialogue response generation. Our approach extends the pointer-generator network to enable the decoder to simultaneously copy tokens from the available set of relevant external knowledge in addition to dialogue context. We demonstrate the effectiveness of our approach through various automatic and human evaluations in comparison with several baselines on the \textsc{ConvAi2} dataset. 
    Furthermore, we conduct diversity, fact inclusion, and error analysis providing further insights into model behaviors.
    In the future, we plan to apply our model to datasets of the same fashion where the dialogue is accompanied by a much larger set of knowledge facts (e.g., Wikipedia articles) \cite{dstc7-track2}. This could be done by adding a retrieval component which identifies a few contextually relevant facts \cite{Ghazvininejad2018Grounded} to be used as input to \textsc{DeepCopy}.
    
    \bibliography{acl2019}
    \bibliographystyle{acl_natbib}

\end{document}